\documentclass{article}

\PassOptionsToPackage{numbers, sort}{natbib}

\usepackage[final,dblblindworkshop]{neurips_2026}
\workshoptitle{Linguistic Principles for Foundation Models (LP4FM)}

\usepackage[utf8]{inputenc}      %
\usepackage[T1]{fontenc}         %
\usepackage{algorithm,algorithmicx}
\usepackage[noend]{algpseudocode}
\usepackage{amsmath,amssymb,amsfonts,amsthm,mathtools}
\usepackage{duckuments}
\usepackage{bm}
\usepackage{hyperref}            %
\usepackage{url}                 %
\usepackage{booktabs}            %
\usepackage[nice]{nicefrac}      %
\usepackage{microtype}           %
\usepackage[dvipsnames]{xcolor}  %
\usepackage{tcolorbox}           %
\usepackage{listings}            %
\usepackage{multirow}            %
\usepackage{colortbl}            %
\usepackage{cprotect}            %
\usepackage{lipsum}
\usepackage{enumitem}
\usepackage[capitalize,noabbrev]{cleveref}
\usepackage{graphicx}
\usepackage{float}
\usepackage[multiple]{footmisc}
\usepackage{caption,subcaption}
\usepackage{textcomp}
\usepackage{stfloats}
\usepackage{longtable,tabularx,booktabs,wrapfig}
\usepackage{makecell}
\usepackage{siunitx}
\usepackage{tabularray}
\tcbuselibrary{breakable,skins}
\usepackage{fancyvrb}
\usepackage{placeins}
\usepackage{amsmath,amsfonts,bm}

\def\eqref#1{equation~\ref{#1}}

\def\1{\bm{1}}

\DeclareMathAlphabet{\mathsfit}{\encodingdefault}{\sfdefault}{m}{sl}
\SetMathAlphabet{\mathsfit}{bold}{\encodingdefault}{\sfdefault}{bx}{n}

\DeclareMathOperator*{\argmax}{arg\,max}

\definecolor{darkblue}{rgb}{0, 0, 0.5}
\definecolor{cadmiumgreen}{rgb}{0.0, 0.42, 0.24}
\definecolor{cornellred}{rgb}{0.7, 0.11, 0.11}
\hypersetup{
  urlcolor=MidnightBlue,
  linkcolor=cornellred,
  citecolor=MidnightBlue,
  colorlinks=true,
}

\title{Representation-Aligned Auxiliary Supervision\\for Language Model Adaptation}

\author{\textbf{Kyuyoung Kim}\textsuperscript{1}\; 
    \textbf{Peiyao Sheng}\textsuperscript{2}\;
    \textbf{Ashwin Hebbar}\textsuperscript{3}\;
    \textbf{Peiyang Xu}\textsuperscript{3}\\
    \textbf{Yunfei Xie}\textsuperscript{4}\;
    \textbf{Kevin Wang}\textsuperscript{5}\;
    \textbf{Rui Xin}\textsuperscript{6}\;
    \textbf{Chen Wei}\textsuperscript{4}\;
    \textbf{Zhangyang Wang}\textsuperscript{5}\\
    \textbf{Jinwoo Shin}\textsuperscript{1}\;
    \textbf{Pramod Viswanath}\textsuperscript{3}\;
    \textbf{Sewoong Oh}\textsuperscript{6} \\
    \textsuperscript{1}KAIST AI\; \textsuperscript{2}Sentient Labs\;
    \textsuperscript{3}Princeton University\; \\ \textsuperscript{4}Rice University\;
    \textsuperscript{5}University of Texas, Austin\; \textsuperscript{6}University of Washington
}

\begin{document}

\maketitle

\begin{abstract}

Language models exhibit strong reasoning capabilities, yet adapting them to structured domains remains challenging and can yield inconsistent outcomes.
We identify \textit{representation compatibility}, the extent to which a model effectively processes a representation for a structured task, as a key factor in adaptation.
We study this in chess, which provides a controlled testbed with precise semantics, computable optimal actions, and multiple state representations, including a symbolic encoding (FEN) and a spatial format (ASCII).
We find that models often process semantically equivalent inputs substantially differently, affecting both learning and generalization.
Building on this observation, we propose \textit{representation-aligned auxiliary supervision}, which uses environment-derived tasks expressed in compatible representations to improve adaptation to structured domains.
Across models and representations, auxiliary supervision consistently improves optimal-move prediction relative to target-only training under identical target data.
Tasks that expose environment dynamics provide larger and most consistent gains than surface-level or static supervision, while remaining competitive with substantially increasing the amount of target-task data.
Moreover, ASCII-trained models transfer more effectively to FEN than FEN-trained models do to ASCII, even surpassing the FEN target-only baseline on FEN evaluation.
The gains also extend beyond optimal-move prediction to open-ended, factually grounded commentary generation.
Overall, our results show that auxiliary supervision in model-compatible representations can enable effective adaptation in structured domains.

\end{abstract}

\section{Introduction}
\label{s:intro}

Language models are increasingly used for prediction and decision-making in structured domains, where an input corresponds to an underlying state ($s$) and the desired output is an action ($a$).
Adapting models to such domains typically relies on supervised fine-tuning (SFT) on task-specific data, optionally followed by reinforcement learning (RL)~\citep{guo2025deepseek,shao2024deepseekmath}.
While this directly supervises the desired state-to-action behavior, obtaining sufficient target-specific supervision can be costly. 
Structured environments, however, often provide additional signals that can be derived automatically, such as state properties, action constraints, and state transitions.
Auxiliary-task learning leverages these signals to provide the model with exposure to the environment structure that may help it learn the target state-to-action mapping~\citep{jaderberg2017unreal}.
We ask when and how such auxiliary supervision can effectively improve adaptation of language models to structured domains.

\begin{figure}[t]
    \centering
    \includegraphics[width=0.95\linewidth]{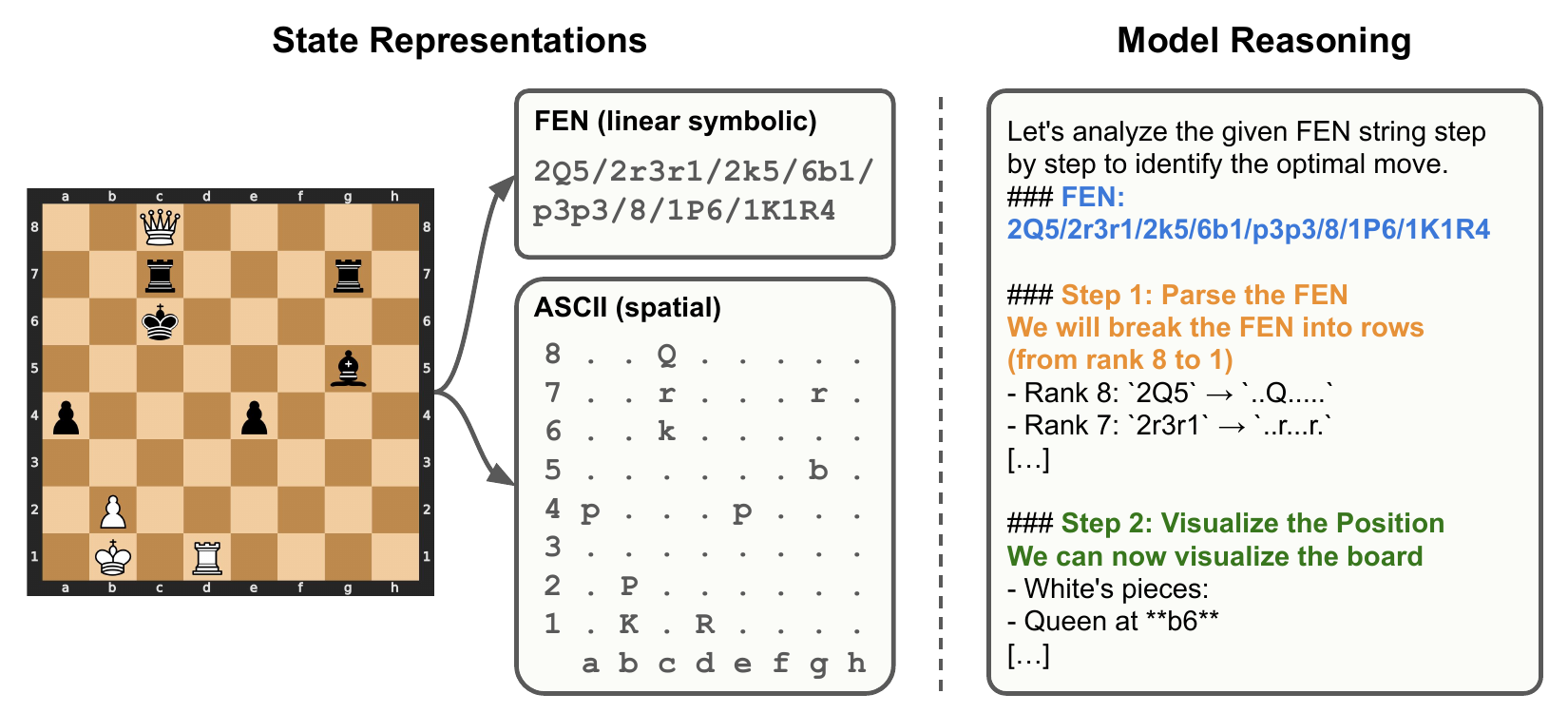}
    \vspace{-5pt}
    \caption{\textbf{Representation compatibility.} Given a chess position in FEN, Qwen3-8B reconstructs an ASCII board before reasoning about the task, illustrating how a model's processing of a structured state can depend substantially on its representation. FEN is a compact linear encoding, while ASCII provides an explicit $8\times8$ spatial representation.
    }
    \label{fig:sample_trace}
    \vspace{-10pt}
\end{figure}

A key question is how the underlying state should be represented when providing such auxiliary learning signals.
In this work, we study \textit{representation compatibility}, the extent to which a model effectively processes a structured state when expressed in a particular representation, as a factor in domain adaptation.
This is motivated by an empirical observation: models may transform inputs into alternative forms prior to reasoning about the target task.
For example, in chess, when given positions in Forsyth--Edwards Notation (FEN), a compact symbolic encoding, Qwen3-8B frequently converts them into ASCII board diagrams and performs reasoning in this spatial format (Figure~\ref{fig:sample_trace}).
Consistent with this behavior, in the task of predicting next states after moves, the model achieves 47.4\% accuracy in ASCII versus 3.9\% in FEN, a twelve-fold gap despite their semantic equivalence (Figure~\ref{fig:baseline-perf}).
This asymmetry suggests that models do not process semantically equivalent inputs uniformly, making representation compatibility an important consideration for adaptation.

Building on this observation, we ask whether auxiliary tasks derived from the environment can improve adaptation to the target state-to-action task and what role representation compatibility plays in the process.
To this end, we introduce a simple training approach that jointly trains on the target objective and environment-derived auxiliary tasks.
These tasks, including predicting static state properties, state transitions, and action constraints, provide additional supervision about environment structure relevant to the target prediction task.

We study these questions in chess, where precise semantics, computable environment structure, and multiple representations enable controlled evaluation of auxiliary supervision and representation choice.
Across our experiments, environment-derived auxiliary tasks consistently improve optimal-move prediction over target-only training.
These gains can often match those obtained by simply increasing the amount of target-task data: although optimal-move accuracy improves with additional examples, environment-derived tasks provide comparable gains without increasing the target-data budget.
The largest gains come from tasks that expose environment dynamics, such as state transitions and action constraints, whereas simpler tasks based on state manipulation or static board properties provide smaller and less consistent improvements.
We further find that the role of representation choice depends on the evaluation setting: its effects are most pronounced for state-understanding tasks and cross-representation transfer, while in-distribution target performance remains relatively robust across representations.
Together, these results suggest that auxiliary supervision provides a complementary learning signal for state-to-action task, and that the representation in which the underlying structure is learned is important for effective adaptation and transfer.

\textbf{Contributions.} In summary, we identify representation compatibility as an important factor in adapting language models to structured domains, and use this insight to develop a simple approach that jointly trains on the target action-prediction objective and environment-derived auxiliary tasks in model-compatible representations.
Our contributions are threefold:
\begin{itemize}[leftmargin=7mm]
\item \textbf{Representation compatibility as a factor in adaptation.} We show that semantically equivalent inputs can differ substantially in how effectively a model learns environment structure, and characterize how representation choice affects performance across tasks (Figure~\ref{fig:baseline-perf}).
\item \textbf{Environment-derived supervision for adaptation.} We introduce a simple joint-training approach that incorporates auxiliary tasks derived from the environment into adaptation, and show that these tasks improve state-to-action prediction beyond target-only training and beyond simply increasing the amount of target data (Table~\ref{table:main-table}, Figure~\ref{fig:main-controls}).
\item \textbf{Analysis of auxiliary supervision and representation.} Through controlled experiments, we identify dynamics-oriented tasks as effective sources of supervision, and show that the effects of representation choice are particularly pronounced for state-understanding tasks and cross-representation transfer (Figure~\ref{fig:task-abl}, Table~\ref{table:commentary-table}).
\end{itemize}

\section{Background and problem setup}
\label{s:background}

We consider fine-tuning a model $\mathcal{M}$ for a target task in which each input corresponds to an underlying \emph{state} $s \in \mathcal{S}$ and the model outputs an \emph{action} $a \in \mathcal{A}$.
Fine-tuning can thus be viewed as learning a conditional distribution over actions given states, $p(a\mid s)$.
For example, in chess, the state is the current board position and the action is the next move.
A \emph{representation} $r \in \mathcal{R}$ defines a mapping $\phi_r(s)$ from a state to an input sequence.
A state may admit multiple semantically equivalent representations that differ substantially in how readily a model can process and reason over them.

\paragraph{Representation compatibility.}
Prior work has observed that reasoning performance can vary substantially across representations that encode the same underlying state.
For example, in code generation, reasoning in Python rather than natural language has been shown to substantially affect performance~\citep{li2024chain,puerto2024code,wen2024unlocking,weir2024learning}.
We use \textit{representation compatibility} to describe how effectively a model processes an underlying state when it is expressed in a particular representation.
Operationally, we assess compatibility by comparing downstream task performance across representations (e.g., Figure~\ref{fig:baseline-perf}) and by examining reasoning behavior of the model (e.g., Figure~\ref{fig:sample_trace}).

In chess reasoning, we observe a pronounced difference between ASCII, a spatially structured representation, and FEN, a linear symbolic encoding (Figure~\ref{fig:baseline-perf}).
This pattern is consistent with prior observations in grid-based environments such as Sokoban, FrozenLake, and Sudoku, where spatial representations facilitate reasoning about environment structures~\cite{chen2025internalizing}.
A complementary qualitative indicator is whether a model transforms one representation into another during reasoning.
Figure~\ref{fig:sample_trace} illustrates this behavior: in an optimal-move prediction task, given a chess position in FEN, Qwen3-8B first converts it into an ASCII representation before reasoning about candidate moves.
The mechanisms underlying these differences, which may reflect factors such as pretraining data or architectural biases, are beyond the scope of this work.
We instead focus on how representation compatibility can be diagnosed and leveraged to improve learning and transfer.

\paragraph{Environment-derived auxiliary tasks.}
In addition to the target objective of predicting the optimal action, we consider auxiliary tasks derived from the underlying environment.
These tasks provide complementary supervision about aspects of the state, action space, and environment structure that are not directly specified by the target objective.
They include tasks involving state properties, legal actions, and next states, covering static properties of the state, aspects of the action space, and transition dynamics.
In particular, next-state prediction models the transition dynamics $p(s' \mid s, a)$, while legal-action prediction characterizes the set of valid actions $\mathcal{A}(s)$.
Because the underlying state can be encoded in multiple ways, these auxiliary tasks can likewise be instantiated using different representations.
We investigate how this choice affects the effectiveness of supervision and its utility to the target task.
Although the formulation is compatible with both supervised and reinforcement learning, we instantiate it using offline data and supervised fine-tuning in this work.

\section{Representation-aligned auxiliary supervision}
\label{s:method}

We propose a simple method for adapting language models to structured state-to-action tasks using environment-derived auxiliary supervision.
The approach uses representation compatibility to determine how this auxiliary supervision is expressed, with the goal of improving target performance.
It consists of three steps: (1) diagnosing representation compatibility, (2) constructing auxiliary tasks from environment structure, and (3) jointly training on the target and auxiliary objectives.

\subsection{Step 1: Diagnosing representational preferences}

Given a set of candidate representations $\mathcal{R}$, we estimate their compatibility with $\mathcal{M}$ by comparing performance across semantically equivalent inputs expressed in different representations.
A natural approach is to evaluate the base model directly on the target task under each representation.
However, when the model lacks sufficient capability, performance may be uniformly low, providing little discriminative signal.
In such cases, we instead use a more tractable \textit{probe task} that preserves the relevant representational structure of the target task for diagnosing compatibility.
For a probe task $\mathcal{T}_{\text{probe}}$ and representation $r$, we estimate compatibility from the model's performance on probe examples $\mathcal{D}_{\text{probe}}$:
\begin{equation*}
    C(\mathcal{M}, r; \mathcal{T}_{\text{probe}}) = f(\mathcal{M}, \phi_r, \mathcal{D}_{\text{probe}}),
\end{equation*}
where $f$ denotes the task-specific evaluation metric.
We use the highest-scoring representation $r^{*}$ as the default representation for auxiliary supervision:
\begin{equation*}
    r^{*} = \argmax_{r \in \mathcal{R}}\, C(\mathcal{M}, r; \mathcal{T}_{\text{probe}}).
\end{equation*}
In addition to task performance, qualitative analysis of reasoning traces provides a complementary signal of compatibility.
For example, if a model consistently transforms inputs into an alternative form before reasoning, this provides behavioral evidence that the alternative representation may be more compatible with its reasoning process (e.g., Figure~\ref{fig:sample_trace}).
Together, probe performance and reasoning analysis provide practical signals for selecting the representation used for auxiliary supervision.

\begin{figure}[h]
    \vspace{-5pt}
    \centering
    \includegraphics[width=0.85\linewidth]{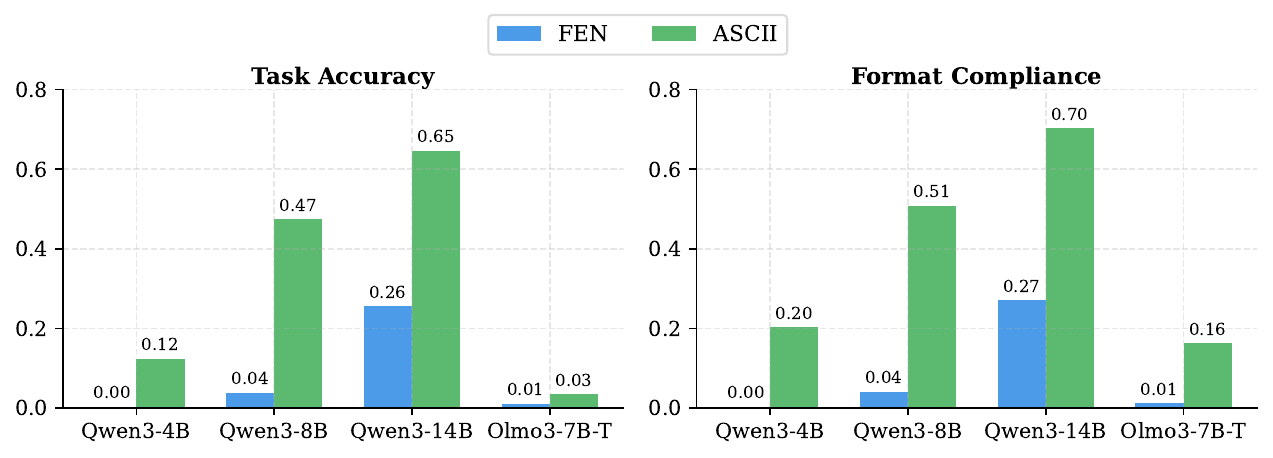}
    \vspace{-5pt}
    \caption{\textbf{Representation compatibility in next-state prediction.} Across model scales and families, ASCII inputs consistently yield higher board accuracy (left) and format compliance (right) than FEN prior to domain adaptation. The gap narrows with model size but remains substantial.}
    \label{fig:baseline-perf}
    \vspace{-5pt}
\end{figure}

As an example, we diagnose compatibility in chess using next-state prediction as a probe, evaluating board accuracy and format compliance under FEN and ASCII (Figure~\ref{fig:baseline-perf}).
Across models, ASCII yields substantially higher accuracy than FEN, providing a practical signal for selecting the representation in which to express auxiliary supervision.
Task accuracy and format compliance are formally defined in Section~\ref{sec:setup}, and task performance after fine-tuning is reported in Table~\ref{tab:reason-len}.

\subsection{Step 2: Constructing environment-derived auxiliary tasks}
\label{sec:worldmodel}

Given a target task $\mathcal T$, we construct auxiliary tasks ${\mathcal T_{\text{aux}}^{(i)}}$ from information that can be computed from the environment.
These tasks provide additional supervision over the representation, underlying state, or environment dynamics without requiring additional target-task annotations.
We consider three types of auxiliary tasks:
\begin{itemize}[leftmargin=7mm]
\item \textbf{Representation-level tasks:} operations over the input representation itself, such as extracting, concatenating, or manipulating components of the representation.
\item \textbf{State-property tasks:} tasks that predict properties of the underlying state independently of a particular action, such as object counts, locations, or occupancy.
\item \textbf{Dynamics tasks:} tasks that characterize how actions interact with the environment, including predicting state transitions and identifying valid actions.
\end{itemize}
We instantiate auxiliary supervision in chess as \textit{surface}, \textit{static}, and \textit{dynamics} tasks, respectively.
Surface tasks operate directly on the board representation, for example by extracting or manipulating ranks.
Static tasks query properties of the board, such as piece counts or material balance.
Dynamics tasks capture the consequences and constraints of actions, including next-state prediction and legal-move prediction.
Each auxiliary task as well as the target task is expressed using the representation $r^*$ selected in Step 1.
This is motivated by the hypothesis that auxiliary supervision is most useful when the representation used to provide it is compatible with the model's processing preferences.
As our experiments in Section~\ref{s:experiments} show, the representation choice is particularly important for state-understanding tasks and cross-representation transfer.

\begin{algorithm}[t]
\caption{Representation-Aligned Auxiliary Supervision}
\label{alg:main}
\begin{algorithmic}[1]

\Require Model $\mathcal{M}$, target task $\mathcal{T}$, auxiliary tasks $\{ \mathcal{T}_{\text{aux}}^{(i)} \}$, candidate representations $\mathcal{R}$, probe task $\mathcal{T}_{\text{probe}}$, probe data $\mathcal{D}_{\text{probe}}$, training data $\mathcal{D}$, loss coefficients $\{\lambda_i\}$
\vspace{0.05in}\hrule\vspace{0.05in}

\Statex \textcolor{gray}{// Step 1: Diagnose representation compatibility}
\For{each $r \in \mathcal{R}$}
    \State $C(\mathcal{M}, r; \mathcal{T}_{\text{probe}}) \gets f(\mathcal{M}, \phi_r, \mathcal{D}_{\text{probe}})$
\EndFor
\State $r^{*} \gets \argmax_{r \in \mathcal{R}} C(\mathcal{M}, r; \mathcal{T}_{\text{probe}})$  \Comment{select the most compatible representation}

\Statex \textcolor{gray}{// Step 2: Instantiate tasks in $r^{*}$}
\For{each $\mathcal{T}_i \in \mathcal{T} \cup \{ \mathcal{T}_{\text{aux}}^{(i)} \}$}
    \State $\mathcal{D}_{\mathcal{T}_i} \gets \phi_{r^{*}}(\mathcal{D}_{\mathcal{T}_i})$  \Comment{express tasks in $r^{*}$}
\EndFor

\Statex \textcolor{gray}{// Step 3: Joint training}
\For{each training step}
    \State Sample batch from $\mathcal{T} \cup \{ \mathcal{T}_{\text{aux}}^{(i)} \}$  \Comment{joint training on the target and auxiliary tasks}
    \State $\mathcal{L} \gets \mathcal{L}(\mathcal{T}) + \sum_i \lambda_i \cdot \mathcal{L}(\mathcal{T}_{\text{aux}}^{(i)})$
    \State Update $\mathcal{M}$ on $\mathcal{L}$
\EndFor

\State \Return $\mathcal{M}$

\end{algorithmic}
\end{algorithm}

\subsection{Step 3: Joint training}

We jointly train the model on the target task $\mathcal{T}$ and auxiliary tasks ${\mathcal{T}_{\text{aux}}^{(i)}}$, with inputs expressed in $r^*$.
Let $\mathcal{L}(\mathcal{T})$ and $\mathcal{L}({\mathcal{T}_{\text{aux}}^{(i)}})$ denote their respective losses.
The training objective is then
\begin{equation*}
\mathcal{L} = \mathcal{L}({\mathcal{T}}) + \sum_i \lambda_i \cdot \mathcal{L}({\mathcal{T}_{\text{aux}}^{(i)}}),
\end{equation*}
where the weights $\lambda_i \geq 0$ control the contribution of each auxiliary task.
The target task remains the primary objective, while the auxiliary losses provide additional learning signals derived from the environment.
We use joint training so that these signals shape the same model used for the target task rather than requiring auxiliary capabilities to be learned independently and composed at inference time.
In our experiments, we compare this joint-training strategy with such sub-skill mixing, where component skills are learned independently and composed at inference time; the latter does not recover the same gains (Table~\ref{tab:skill-mix}).

\section{Experiments}
\label{s:experiments}

\subsection{Experimental setup}
\label{sec:setup}

We study domain adaptation in chess, where board states have precise semantics, environment-derived supervision can be computed automatically, and the same state admits multiple representations.
This setting allows us to control both the type of auxiliary supervision and its representation while evaluating performance on state-to-action prediction.

\paragraph{Representations.}
We consider two representations of chess positions: Forsyth--Edwards Notation (FEN) and ASCII board diagrams.
FEN encodes the board as a linear symbolic sequence, where each rank is described by characters for piece types and digits indicating consecutive empty squares.
In contrast, ASCII represents the board as an $8\times8$ spatial grid with piece symbols placed at their corresponding locations and empty squares denoted by dots.
The two representations are semantically equivalent and admit exact transformations.

\paragraph{Tasks.}

Our primary target task is optimal-move prediction: given a board state, the model predicts the best move as evaluated by a chess engine.
This task requires reasoning about the current board state, evaluating candidate moves, and identifying the strongest one.
We consider three types of environment-derived auxiliary supervision.
\emph{Surface} tasks operate on the representation itself through operations such as extracting, concatenating, or reversing portions of the board encoding.
\emph{Static} tasks predict properties of the board state, such as material counts, piece locations, or square occupancy.
\emph{Dynamics} supervision captures relationships between states and actions through next-state prediction and legal-move prediction.
We also consider simply providing additional optimal-move training data, which serves as a control for training-data volume.
All tasks use chain-of-thought prompting, with reasoning enclosed in \texttt{<think>} tags and final answers in \texttt{<answer>} tags.

\paragraph{Data.}
We construct datasets for optimal-move prediction and environment-derived auxiliary tasks using the Lichess Evaluation Database, which contains diverse chess positions annotated with Stockfish analysis.
The primary optimal-move dataset contains 100k board-move pairs, augmented with reasoning traces derived from engine evaluations.
For the auxiliary tasks, we construct 200k examples for each supervision condition, with task-specific details described in Appendix~\ref{app:data-details}.
Evaluation is performed on a held-out set of 5k positions disjoint from the training data.

\paragraph{Training and evaluation.}
Our main experiments use Qwen3-8B and Qwen3-14B~\citep{yang2025qwen3} as the base models.
We additionally evaluate Olmo-3-7B-Think~\citep{olmo2025olmo} to assess generalization across model families.
All models are trained via supervised fine-tuning (SFT) using the objective described in Section~\ref{s:method}.
Further training details are provided in Appendix~\ref{app:train-details}.

We report exact-match (EM) accuracy, the fraction of predictions that exactly match the reference.
For next-state prediction, we additionally report board accuracy (BA), which scores predictions rank-by-rank and assigns $k/8$ for $k$ correctly predicted ranks.
For legal-move prediction, we use binary correctness: 1 if the predicted move is legal and 0 otherwise.
Format compliance measures the fraction of responses that conform to the expected output format and is reported where relevant as a measure of instruction-following capability.

\subsection{Environment-derived supervision for target adaptation}
\label{sec:results}

\begin{table}[t]
    \centering
    \small  %
    \caption{\textbf{Environment-derived supervision improves optimal-move prediction.}
    OM trains on optimal-move prediction only; OM + dynamics additionally includes next-state and legal-move supervision.
    Training representations are shown in parentheses; FEN and ASCII columns report accuracy under each representation.
    Shading indicates in-distribution performance.}
    \label{table:main-table}
    \vspace{2pt}
    \begin{tabular}{l*{6}{c}}
    \toprule[1pt]
    \multirow{2.5}{*}{\textbf{Training}}
        & \multicolumn{2}{c}{\textbf{{Qwen3-8B}}}
        & \multicolumn{2}{c}{\textbf{{Qwen3-14B}}}
        & \multicolumn{2}{c}{\textbf{Olmo3-7B-T}} \\
    \cmidrule(lr){2-3} \cmidrule(lr){4-5} \cmidrule(lr){6-7}
     & FEN & ASCII & FEN & ASCII & FEN & ASCII \\
    \midrule[0.75pt]
    Baseline           & 0.0   & 0.001
                       & 0.020 & 0.041
                       & 0.0   & 0.0   \\
    \midrule[0.25pt]
    OM only (FEN)         & \cellcolor{gray!15}0.181 & 0.146
                          & \cellcolor{gray!15}0.199 & 0.161
                          & \cellcolor{gray!15}0.172 & 0.152 \\
    OM + dynamics (FEN)   & \cellcolor{gray!15}\textbf{0.192} & 0.153
                          & \cellcolor{gray!15}\textbf{0.223} & 0.166
                          & \cellcolor{gray!15}\textbf{0.185} & 0.077 \\
    \midrule[0.25pt]
    OM only (ASCII)       & 0.168 & \cellcolor{gray!15}0.181
                          & 0.194 & \cellcolor{gray!15}0.206
                          & 0.157 & \cellcolor{gray!15}0.176 \\
    OM + dynamics (ASCII) & 0.173 & \cellcolor{gray!15}\textbf{0.201}
                          & 0.202 & \cellcolor{gray!15}\textbf{0.226}
                          & 0.185 & \cellcolor{gray!15}\textbf{0.199} \\
    \bottomrule[1pt]
    \end{tabular}
    \vspace{-10pt}
\end{table}

We first ask whether environment-derived auxiliary supervision improves performance on the target task.
Table~\ref{table:main-table} compares optimal-move prediction with and without dynamics supervision, reporting mean accuracy across three training seeds.
For the evaluated models, adding this auxiliary signal consistently improves optimal-move accuracy.
For example, with ASCII training, the additional supervision improves Qwen3-14B accuracy from 0.206 to 0.226 (+9.7\%), while with FEN training it improves accuracy from 0.199 to 0.223 (+12.1\%).
These improvements are obtained with the same 100k optimal-move training examples, indicating that environment-derived supervision provides a useful learning signal beyond the target objective.

\begin{figure}[t]
    \centering
    \includegraphics[width=0.85\linewidth]{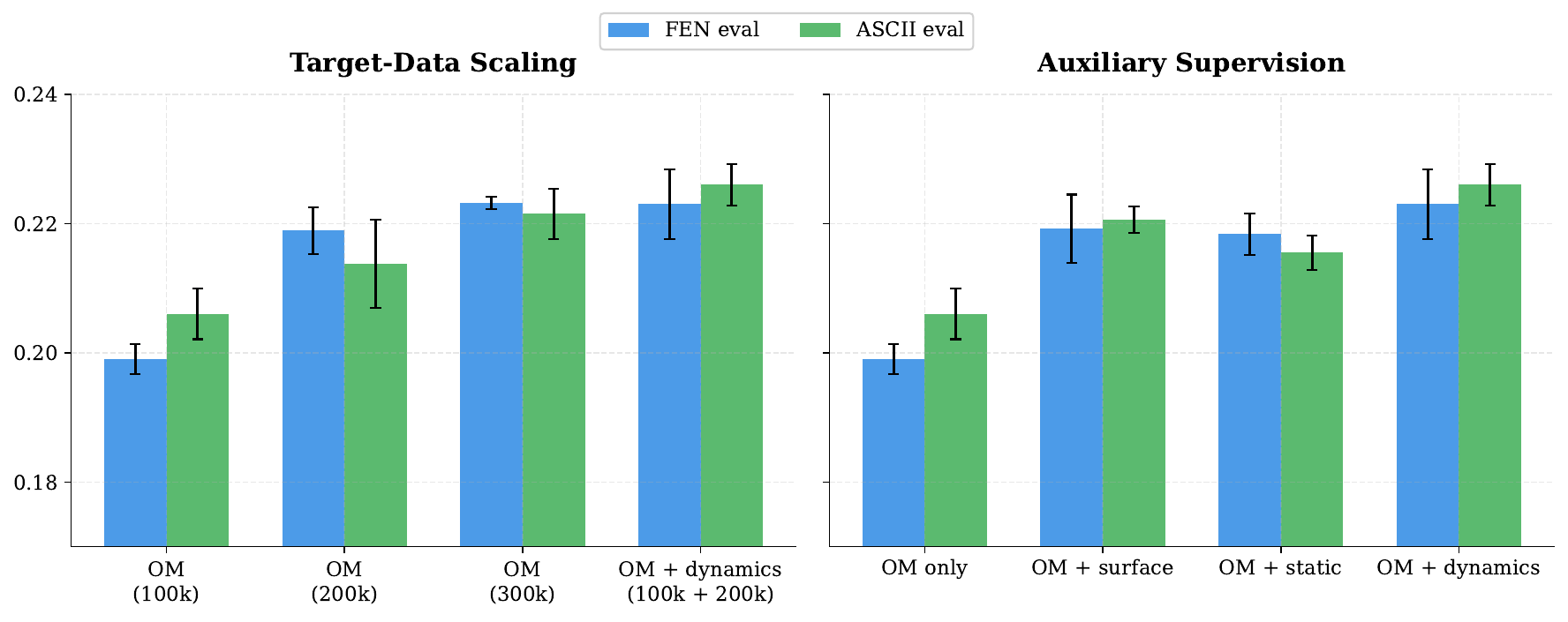}
    \vspace{-5pt}
    \caption{\textbf{Environment-derived supervision improves beyond target-data scaling and matched-data controls.}
    (a) Optimal-move accuracy as the amount of target-task training data increases, compared with 100k target-task examples plus 200k environment-derived dynamics examples.
    (b) Optimal-move accuracy under matched-data auxiliary supervision conditions.
    All results use Qwen3-14B, with models trained and evaluated in either FEN or ASCII.
    }
    \label{fig:main-controls}
    \vspace{-5pt}
\end{figure}

The gains in Table~\ref{table:main-table} could in principle arise simply from exposing the model to more training data, rather than from the information provided by the auxiliary tasks.
We therefore compare auxiliary supervision with increasing the amount of target-task data.
As shown in Figure~\ref{fig:main-controls}a, optimal-move performance in ASCII increases from 0.206 with 100k target examples to 0.214 with 200k and 0.222 with 300k.
In contrast, combining the original 100k target examples with 200k additional dynamics examples yields an accuracy of 0.226, comparable to that achieved by training on 300k target examples.
A similar pattern is observed for FEN.
Thus, environment-derived supervision can provide gains comparable to those obtained by substantially increasing the amount of target-task data.

We next examine which properties of the auxiliary supervision contribute to these gains.
Figure~\ref{fig:main-controls}b compares supervision conditions under a matched budget of 200k additional examples.
Surface tasks and static state-property tasks produce smaller and less consistent improvements, whereas dynamics supervision yields the largest gain. %
This pattern suggests that the most useful auxiliary supervision is not merely additional multitask training or exposure to static properties of the state, but supervision that captures relationships between states and actions through transitions and action constraints.

Together, these results show that environment-derived auxiliary supervision can improve adaptation to structured state-to-action tasks beyond target-only training and beyond simply increasing target-task data.
Moreover, the matched-data comparisons indicate that the content of the auxiliary supervision matters: supervision that exposes environment dynamics provides particularly strong gains.

\subsection{Representation compatibility and transfer}
\label{sec:representation}

We next examine how the representation used for auxiliary supervision affects what the model learns and how that knowledge transfers.
Our probe experiments demonstrate a notable gap for next-state prediction, with models performing substantially better in ASCII than in FEN (Figure~\ref{fig:baseline-perf}).
For optimal-move prediction, however, the effect is comparatively modest.
As shown in Table~\ref{table:main-table}, models trained and evaluated in their respective representations achieve broadly similar performance, with the differences between FEN and ASCII smaller than those observed in state understanding.
Thus, representation compatibility does not necessarily translate into large in-distribution gains on the target task, particularly when it can be learned reasonably well from either representation.

The effect becomes more pronounced when evaluating transfer across representations.
Models trained with dynamics supervision in ASCII consistently transfer more effectively to FEN than models trained in FEN transfer to ASCII.
For example, with Qwen3-14B, OM + dynamics (ASCII) achieves 0.202 when evaluated in FEN, compared with 0.166 for OM + dynamics (FEN) evaluated in ASCII.
Notably, the ASCII-trained model exceeds 0.199 achieved by OM-only training in FEN.
A similar pattern is observed for Olmo3-7B: OM + dynamics (ASCII) achieves 0.185 when evaluated in FEN, matching the in-distribution performance of FEN, whereas OM + dynamics (FEN) evaluated in ASCII drops substantially to 0.077.
This asymmetry suggests that the representation used for auxiliary training affects how effectively the resulting knowledge transfers across semantically equivalent forms.

\begin{table}[ht]
    \vspace{-5pt}
    \centering
    \small  %
    \caption{\textbf{Representation on performance and reasoning efficiency.} Each row corresponds to Qwen3-8B fine-tuned on a representation and reasoning length combination. \textbf{Avg.\ Len.} denotes average response length. \textbf{Fmt}: format compliance; \textbf{EM}: exact match; \textbf{BA}: board accuracy.}
    \label{tab:reason-len}
    \vspace{2pt}
    \setlength{\tabcolsep}{5pt}
    \begin{tabular}{l*{4}{c}}
    \toprule[1pt]
    \textbf{Training Data} & \textbf{Avg.\ Len.} & \textbf{Fmt} & \textbf{EM} & \textbf{BA} \\
    \midrule[0.75pt]
    FEN$\to$FEN (short)             & 267  & 1.000 & 0.539 & 0.916 \\
    FEN$\to$FEN (long)              & 622  & 1.000 & \textbf{0.613} & \textbf{0.930} \\
    \midrule[0.25pt]
    ASCII$\to$ASCII (short)         & 258  & 1.000 & \textbf{0.934} & \textbf{0.988} \\
    ASCII$\to$ASCII (long)          & 681  & 0.999 & 0.713 & 0.942 \\
    \midrule[0.25pt]
    FEN$\to$ASCII$\to$FEN (short)   & 276  & 1.000 & 0.824 & 0.971 \\
    FEN$\to$ASCII$\to$FEN (long)    & 664  & 1.000 & \textbf{0.912} & \textbf{0.981} \\
    \bottomrule[1pt]
    \end{tabular}
    \vspace{-10pt}
\end{table}

The contrast between representations is particularly clear for state-understanding tasks.
After fine-tuning Qwen3-8B, ASCII$\to$ASCII achieves 93.4\% exact match compared with 61.3\% for FEN$\to$FEN (Table~\ref{tab:reason-len}), consistent with the gap observed in the base-model probe. %
Together with the results above, this suggests that representation choice has a stronger effect on learning and transferring structured state knowledge than on in-distribution target-task performance.

\subsection{Additional analyses}

\paragraph{Representation compatibility and reasoning length.}
In next-state prediction, representation compatibility also changes how additional reasoning affects learning: longer reasoning improves performance under FEN but degrades performance under ASCII (Table~\ref{tab:reason-len}).
Moreover, using ASCII as an intermediate representation in the FEN$\to$ASCII$\to$FEN setting substantially improves performance over direct FEN$\to$FEN, despite identical input and output formats.
These results suggest that representation compatibility affects both the amount of reasoning that benefits learning and the usefulness of intermediate representations for structured state transformations.

\begin{figure}[ht]
    \vspace{-5pt}
    \centering
    \includegraphics[width=0.85\linewidth]{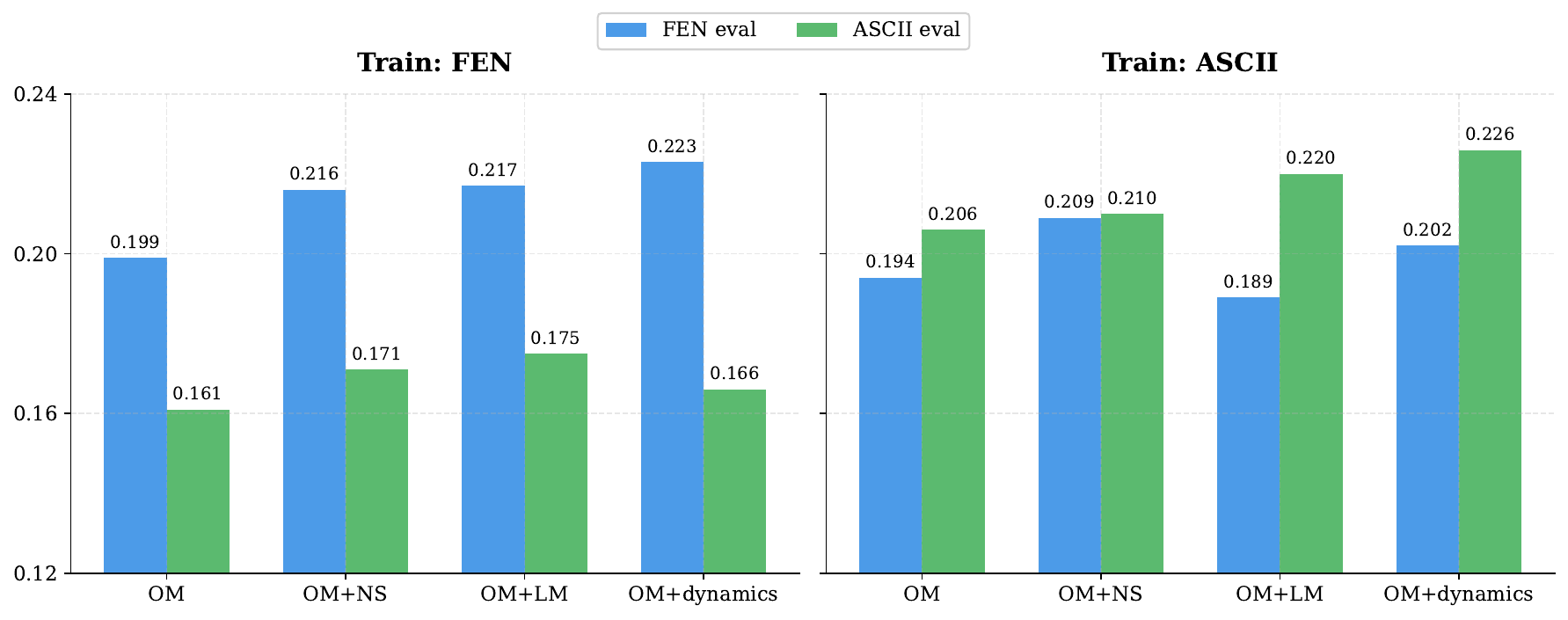}
    \vspace{-5pt}
    \caption{
    \textbf{Contribution of environment-derived auxiliary tasks.}
    Optimal-move accuracy for models trained with different combinations of tasks.
    The tasks improve over target-only training, while their combination (OM + dynamics) yields the largest gains across the two training representations.
    }
    \label{fig:task-abl}
    \vspace{-10pt}
\end{figure}

\paragraph{Auxiliary task composition.}
We analyze the contribution of individual auxiliary tasks by augmenting optimal-move prediction (OM) with next-state prediction (NS), legal-move prediction (LM), and their combination (dynamics).
Figure~\ref{fig:task-abl} reports results for the 14B model.
Across both training representations, NS and LM each provide modest gains over target-only training.
However, combining the two tasks yields larger and more consistent improvements, particularly under ASCII training, which achieves the highest in-distribution performance.
These results suggest that NS and LM provide complementary supervision about environment dynamics and action constraints, and that jointly exposing the model to both aspects is more effective than either task alone.

\paragraph{Auxiliary supervision on open-ended generation.}
We additionally evaluate whether the benefits of environment-derived auxiliary supervision extend beyond move prediction to open-ended generation.
Given a board state and a move, the model generates instructive commentary, which we evaluate for factual correctness using a tool-augmented judge.
Table~\ref{table:commentary-table} reports results under different auxiliary warm-up conditions and commentary representations.
Each row specifies the model initialization: the base model, no auxiliary warm-up, or a model first trained on optimal-move prediction and dynamics supervision in FEN or ASCII.
The columns then indicate the representation used for commentary fine-tuning and evaluation.
See Appendix~\ref{app:commentary-gen} for more details.

\begin{table}[t]
    \vspace{-5pt}
    \centering
    \small
    \caption{\textbf{Auxiliary supervision on commentary generation.}
    Claim accuracy and error rate for Qwen3-8B across auxiliary warm-up conditions.
    Columns indicate the representation used for commentary training and evaluation; rows indicate the preceding training or auxiliary warm-up condition.
    For the OM + dynamics rows, the auxiliary model is trained in the representation shown in parentheses.
    Shading indicates matched auxiliary and commentary representations.
    }
    \label{table:commentary-table}
    \vspace{2pt}
    \begin{tabular}{l*{4}{c}}
    \toprule[1pt]
    \multirow{2.5}{*}{\textbf{Initialization}}
        & \multicolumn{2}{c}{\textbf{FEN}}
        & \multicolumn{2}{c}{\textbf{ASCII}} \\
    \cmidrule(lr){2-3} \cmidrule(lr){4-5}
     & Claim Acc. $\uparrow$ & Error Rate $\downarrow$
     & Claim Acc. $\uparrow$ & Error Rate $\downarrow$ \\
    \midrule[0.75pt]
    Baseline (Qwen3-8B)  & 20.1                    & 61.5
                         & 17.1                    & 56.6     \\
    \midrule[0.25pt]
    No auxiliary supervision  & 33.7                    & 39.0
                              & 44.0                    & 32.2     \\
    \midrule[0.25pt]
    OM + dynamics (FEN)      & \cellcolor{gray!15}50.7 (+17.0)   & \cellcolor{gray!15}{23.2} (-15.8)
                             & 41.7  (-2.3)                      & 35.7 (+3.5)    \\
    OM + dynamics (ASCII)    & 44.0  (+10.3)                     & 36.3 (-2.7)
                             & \cellcolor{gray!15}{51.0} (+7.0)  & \cellcolor{gray!15}27.9 (-4.3) \\
    \bottomrule[1pt]
    \end{tabular}
    \vspace{-10pt}
\end{table}

Without auxiliary warm-up, training and evaluating commentary generation in ASCII outperforms FEN by 10.3 percentage points in claim accuracy, consistent with the model's preference for spatially structured inputs.
Auxiliary supervision further improves commentary generation when the warm-up and commentary representations are aligned.
Starting from the OM + dynamics (ASCII) checkpoint and training commentary generation in ASCII yields 51.0\% claim accuracy and reduces the error rate to 27.9\%.
The corresponding FEN-aligned condition achieves 50.7\% claim accuracy and a 23.2\% error rate.
We also observe stronger cross-representation transfer from ASCII-aligned auxiliary supervision: compared with no warm-up, ASCII warm-up improves claim accuracy by 10.3 percentage points when commentary is trained and evaluated in FEN, whereas FEN warm-up does not improve performance in ASCII.
Overall, these results show that environment-derived supervision also benefit open-ended, factually grounded generation, while the representation affects transfer.

\section{Related work}
\label{s:related}

Chess is a common testbed for studying reasoning in language models due to its precise semantics and computable ground truth.
Prior work has studied state tracking~\citep{toshniwal2022chess}, policy-language training~\citep{feng2023chessgpt}, engine distillation~\citep{ruoss2024grandmaster}, strategy-aware fine-tuning~\citep{wang2024mate}, and commentary generation~\citep{kim2025bridging}, as well as strategic reasoning with RL~\citep{hwang2025can,kim2026correct} and planning with learned world models~\citep{zhang2024mastering}.
We instead use chess as a controlled setting for studying how representation choice affects domain adaptation.

Model performance is known to depend strongly on input representation.
For example, expressing reasoning as code rather than natural language improves accuracy across a range of tasks~\citep{gao2023pal,li2024chain,puerto2024code,wen2024unlocking,weir2024learning}, and structured formats can elicit stronger decision-making~\citep{hu2025define}.
These findings have primarily been exploited through inference-time prompting or representation design.
We instead treat representation compatibility as a training-time consideration, using it to select how environment-derived auxiliary supervision is expressed.

Our work also relates to auxiliary-task learning.
In RL, environment-derived supervision has been widely used to accelerate learning by developing richer representations~\citep{jaderberg2017unreal,ha2018world,hafner2023dreamerv3}.
We study a complementary question in language model adaptation: not only which environment-derived signals to supervise, but also which representation should be used to express them.

\section{Conclusion}
\label{s:conclusion}

We study adaptation of language models to structured domains through the lens of representation compatibility.
Our results show that environment-derived auxiliary supervision can provide a complementary learning signal beyond direct supervision of the target task, but its effectiveness depends on both the structure being learned and how that structure is represented.
In controlled experiments in chess, we find that dynamics-oriented supervision, such as learning state transitions and action constraints, provides particularly strong gains, while representation choice has particularly pronounced effects on state understanding and cross-representation transfer.
We also find that these benefits are not fully recovered by learning auxiliary skills independently and composing them post hoc, suggesting that jointly learning environment structure and the target behavior can be important for effective adaptation.
Overall, our findings highlight representation as an important design consideration for leveraging environment structure in language model adaptation to structured domains.

\section*{Acknowledgments}
This research was supported in part by the Institute for Information \& Communications Technology Planning \& Evaluation (IITP) grant funded by the Korean government (MSIT) (RS-2019-II190075, Artificial Intelligence Graduate School Support Program (KAIST)), by Army Research Office (ARO) Award  W911NF2310147, and by NSF awards CNS-2112471, IIS-2229876, CCF-2505865, SaTC-2325478 and DMS-2502281.
The authors thank Sentient Labs for generously providing the compute resources used in this work.

\bibliographystyle{unsrtnat}
\bibliography{neurips_2026}

\newpage
\appendix

\section{Experimental details}
\label{app:exp-details}

\subsection{Data construction}
\label{app:data-details}

We sample data from the Lichess Evaluations Database\footnote{\url{https://database.lichess.org}}, a large collection of chess positions analyzed by the open-source chess engine Stockfish.
Each entry contains a position in FEN, candidate moves with centipawn scores (measuring positional advantage), mate evaluations when applicable, and a principal variation (PV) corresponding to the engine's predicted best continuation.
For each task and representation, we construct examples consisting of a reference answer and a synthetic reasoning trace tailored to the task.
For optimal-move prediction, the traces reason over Stockfish's top candidate moves.
For auxiliary tasks, the traces focus on the aspects of the board representation relevant to producing the correct answer.

Surface tasks operate purely on the board representation.
Given a board, we sample one of five templates: extracting a specified rank verbatim; counting occurrences of a specified character within a rank; concatenating two specified ranks; identifying the lexicographically greatest rank when ranks are treated as strings; or reversing the character order of a specified rank.
These tasks can be solved by manipulating the board string as text and require no chess-specific knowledge, such as piece identity, move legality, or board geometry.

Static tasks query factual properties of the position without involving transition dynamics.
We sample one of five templates: counting a given piece type for a given side; computing the material balance using standard $1, 3, 3, 5, 9$ piece values while ignoring kings; listing all pieces for one side; counting the total number of pieces including kings; or determining which side has more heavy pieces (queens and rooks).
These tasks require interpreting the board state but do not require reasoning about how the position may evolve.

Legal-move prediction requires reasoning about the rules that determine valid chess moves.
Although chess positions follow well-defined rules, some positions involve additional constraints, including moves that expose the king to check, pinned pieces, check and double-check situations, pawn promotion, and en passant.
We therefore oversample positions exhibiting these conditions to ensure adequate coverage of such edge cases for training and evaluation.

\subsection{Training details}
\label{app:train-details}

\begin{table}[h]
    \centering
    \vspace{-5pt}
    \caption{\textbf{Supervised fine-tuning hyperparameters.}}
    \label{tab:hparams-sft}
    \vspace{2pt}
    \setlength{\tabcolsep}{8pt}
    \begin{tabular}{lll}
    \toprule
     & \textbf{Hyperparameter} & \textbf{Value} \\
    \midrule[0.75pt]
    \textbf{Training} & Epochs & 3 \\
                      & Batch size & 256 \\
                      & Gradient clipping & 1.0 \\
                      & Precision & bf16 \\
    \midrule[0.25pt]
    \textbf{Optimization} & Learning rate & 1e-5 \\
                          & LR scheduler & cosine \\
                          & Warmup ratio & 0.1 \\
                          & Weight decay & 0.01 \\
                          & Loss coefficients $\lambda_i$ for all $i$ & 1.0 \\
    \bottomrule[1pt]
    \end{tabular}
    \vspace{-5pt}
\end{table}

Experiments are conducted primarily on NVIDIA H100 GPUs.
Training time varies across different data mixtures, but most experiments complete within one day on 4 H100 GPUs.
All models are trained with the AdamW optimizer~\citep{loshchilov2019decoupled} using $\beta_1 = 0.9$ and $\beta_2 = 0.95$.
FlashAttention 2~\citep{dao2023flashattention} is enabled for all runs.
Table~\ref{tab:hparams-sft} summarizes the hyperparameters used in training.

\section{Additional experimental results}
\label{app:more-results}

\subsection{Win probability loss evaluation}

Table~\ref{table:main-table} reports optimal-move accuracy, which treats all non-optimal moves equally.
However, chess positions often admit multiple moves with similar quality, making exact-match accuracy a coarse measure of move strength.
To complement this evaluation, we measure \emph{win probability loss} (WPL), defined as the absolute difference in win probability between the optimal Stockfish move and the model-selected move:
\[
\text{WPL} = \left| \text{WinProb}(\text{Stockfish move}) - \text{WinProb}(\text{Model move}) \right|.
\]
We convert engine centipawn (cp) scores to win probabilities using the standard Lichess formula\footnote{\url{https://lichess.org/page/accuracy}}:
\[
\text{WinProb} = 50 + 50 \left( \frac{2}{1 + e^{-0.00368208 \cdot \text{cp}}} - 1 \right).
\]

\begin{table}[t]
    \centering
    \small
    \caption{\textbf{Comparison of move quality in win probability loss (Qwen3-8B).} We report mean and median win probability loss (WPL), as a softer measure of move quality based on changes in engine-estimated win probability. Shading indicates in-distribution performance.}
    \label{table:wp_loss}
    \vspace{2pt}
    \begin{tabular}{lcccc}
    \toprule[1pt]
    \multirow{2.5}{*}{\textbf{Training}}
        & \multicolumn{2}{c}{\textbf{Mean WPL $\downarrow$}}
        & \multicolumn{2}{c}{\textbf{Median WPL $\downarrow$}} \\
    \cmidrule(lr){2-3} \cmidrule(lr){4-5}
     & FEN & ASCII & FEN & ASCII \\
    \midrule[0.75pt]

    OM only (FEN)
        & \cellcolor{gray!15}21.8 {\footnotesize ($\pm$27.2)} & 24.8 {\footnotesize ($\pm$27.2)}
        & \cellcolor{gray!15}$7.8$ & $12.9$ \\
    OM + dynamics (FEN)
        & \cellcolor{gray!15}\textbf{21.0 {\footnotesize ($\pm$26.7)}} & 23.8 {\footnotesize ($\pm$26.8)}
        & \cellcolor{gray!15}\textbf{7.1} & 11.1 \\

    \midrule[0.25pt]

    OM only (ASCII)
        & 22.6 {\footnotesize ($\pm$26.2)} & \cellcolor{gray!15}22.0 {\footnotesize ($\pm$27.2)}
        & 9.4 & \cellcolor{gray!15}8.0 \\
    OM + dynamics (ASCII)
        & 22.5 {\footnotesize ($\pm$26.5)} & \cellcolor{gray!15}\textbf{20.9 {\footnotesize ($\pm$27.0)}}
        & 10.0 & \cellcolor{gray!15}\textbf{6.4} \\

    \bottomrule[1pt]
    \end{tabular}
    \vspace{-5pt}
\end{table}

Table~\ref{table:wp_loss} shows that auxiliary dynamics supervision reduces both mean and median WPL in-distribution for both FEN and ASCII training.
The improvements are also reflected in cross-representation evaluation: dynamics supervision in ASCII achieves lower WPL when evaluated in FEN than dynamics supervision in FEN achieves when evaluated in ASCII, consistent with the stronger cross-representation transfer observed in Table~\ref{table:main-table}.
Although the reductions in mean WPL are modest, the lower median WPL provides additional evidence that auxiliary supervision improves the general quality of the model's selected move beyond exact optimal-move accuracy.

\subsection{Mixture of representations}

We examine whether training on multiple representations can improve performance across evaluation settings, and whether exposure to multiple representations alone induces cross-representation reasoning.
To evaluate this, we train Qwen3-8B on three variants of next-state prediction: FEN$\to$FEN, ASCII$\to$ASCII, and FEN$\to$ASCII$\to$FEN (F2A2F), using the best-performing reasoning length for each setting identified in earlier experiments.
We additionally consider explicit representation-mapping tasks to provide direct supervision for conversions between FEN and ASCII.
The resulting training mixtures are:
\begin{itemize}[leftmargin=7mm]
    \item \textbf{FEN + ASCII (100k):} 50k examples of each single-representation task.
    \item \textbf{FEN + ASCII + F2A2F (150k):} additionally includes 50k mixed-representation examples.
    \item \textbf{FEN + ASCII + F2A2F + Conv (250k):} additionally includes 50k examples for each FEN$\leftrightarrow$ASCII conversion direction.
\end{itemize}

\begin{table}[t]
    \centering
    \small
    \caption{\textbf{Representation mixture and learning.} Performance on next-state prediction under different data compositions. \textbf{Fmt}: format compliance; \textbf{EM}: exact match accuracy; \textbf{BA}: board accuracy.}
    \label{tab:mix-core}
    \vspace{2pt}
    \setlength{\tabcolsep}{4pt}
    \begin{tabular}{l*{9}{c}}
    \toprule[1pt]
    \multirow{2.5}{*}{\textbf{Training}} & \multicolumn{3}{c}{\textbf{FEN$\to$FEN}} & \multicolumn{3}{c}{\textbf{ASCII$\to$ASCII}} & \multicolumn{3}{c}{\textbf{FEN$\to$ASCII$\to$FEN}} \\
    \cmidrule(lr){2-4} \cmidrule(lr){5-7} \cmidrule(lr){8-10}
     & Fmt & EM & BA & Fmt & EM & BA & Fmt & EM & BA \\
    \midrule[0.75pt]
    Best Baseline                  & 1.000 & 0.613 & 0.930 & 1.000 & 0.934 & 0.988 & 1.000 & 0.912 & 0.981 \\
    \midrule[0.25pt]
    FEN + ASCII                    & 1.000 & 0.741 & 0.952 & 1.000 & 0.908 & 0.983 & 1.000 & 0.016 & 0.647 \\
    FEN + ASCII + F2A2F            & 1.000 & 0.805 & 0.965 & 1.000 & 0.937 & 0.989 & 1.000 & 0.917 & 0.982 \\
    FEN + ASCII + F2A2F + Conv     & 1.000 & \textbf{0.860} & \textbf{0.974} & 1.000 & \textbf{0.953} & \textbf{0.991} & 1.000 & \textbf{0.939} & \textbf{0.987} \\
    \bottomrule[1pt]
    \end{tabular}
    \vspace{-5pt}
\end{table}

Table~\ref{tab:mix-core} shows that combining representations improves next-state prediction on both FEN and ASCII relative to the corresponding single-representation baselines.
However, simply mixing FEN and ASCII training data does not induce cross-representation reasoning: the FEN + ASCII model achieves only 0.016 EM on FEN$\to$ASCII$\to$FEN, despite strong performance on both single-representation tasks.
Adding explicit FEN$\to$ASCII$\to$FEN training substantially improves performance on the mixed-representation task, while also improving the single-representation tasks.
Finally, adding explicit conversion supervision further improves performance across the evaluation settings.
These results suggest that exposure to multiple representations alone is insufficient to reliably induce their coordinated use; explicit supervision for the desired cross-representation behavior is substantially more effective.

\subsection{Limits of sub-skill composition}

Our framework jointly trains on the target and auxiliary tasks in an aligned representation.
We ask whether the same behavior can instead be obtained by training the constituent skills separately and relying on the model to compose them at inference time.
To study this, we decompose the FEN$\to$ASCII$\to$FEN task into four sub-skills: (S1) FEN-to-ASCII conversion, (S2) ASCII-to-FEN conversion, (S3) ASCII-based reasoning, and (S4) FEN-based reasoning.
We train Qwen3-8B on mixtures of these sub-skills without exposing it to the full task, and evaluate whether it can compose them at inference time.

\begin{table}[h]
    \vspace{-5pt}
    \centering
    \caption{\textbf{Limits of sub-skill composition.} Left: models trained on constituent reasoning skills fail to compose them into the FEN$\to$ASCII$\to$FEN pipeline. Right: adding FEN-to-ASCII conversion to FEN-based reasoning does not induce the use of ASCII as an intermediate representation. Source indicates whether the constituent tasks use the same set of positions (Same) or disjoint sets (Diff).}
    \label{tab:skill-mix}
    \vspace{2pt}
    \begin{minipage}{0.48\linewidth}
      \centering
      \small  %
      \setlength{\tabcolsep}{3pt}
      \begin{tabular}{l*{5}{c}}
        \toprule[1pt]
        \textbf{Training Data} & \textbf{Source} & \textbf{Fmt} & \textbf{EM} & \textbf{BA} \\
        \midrule[0.75pt]
        FEN$\to$ASCII$\to$FEN & -- & 1.000 & 0.912 & 0.981 \\
        \midrule[0.25pt]
        S1 + S3 + S2 & Same & 0.975 & 0.003 & 0.644 \\
        S1 + S3 + S2 & Diff & 0.972 & 0.004 & 0.633 \\
        \bottomrule[1pt]
      \end{tabular}
    \end{minipage}
    \hfill
    \begin{minipage}{0.48\linewidth}
      \centering
      \small  %
      \setlength{\tabcolsep}{4pt}
      \begin{tabular}{l*{5}{c}}
        \toprule
        \textbf{Training Data} & \textbf{Source} & \textbf{Fmt} & \textbf{EM} & \textbf{BA} \\
        \midrule[0.75pt]
        ASCII$\to$ASCII & -- & 1.000 & 0.934 & 0.988 \\
        \midrule[0.25pt]
        S1 + S4 & Same & 0.985 & 0.055 & 0.743 \\
        S1 + S4 & Diff & 0.983 & 0.060 & 0.740 \\
        \bottomrule[1pt]
      \end{tabular}
    \end{minipage}
    \vspace{-5pt}
\end{table}

Table~\ref{tab:skill-mix} shows that the constituent skills do not compose effectively at inference time.
Models trained on S1 + S3 + S2 achieve near-zero exact match on the full FEN$\to$ASCII$\to$FEN task, despite being trained on each component of the pipeline.
Similarly, augmenting FEN-based reasoning with conversion skills does not induce the use of ASCII as an intermediate representation.
These results suggest that, at this scale, independently trained constituent skills are insufficient to reliably induce the desired reasoning structure.

\section{Commentary generation and evaluation}
\label{app:commentary-gen}

\subsection{Data curation}
We collect annotated games from classical chess books and online forums, including Chess Publishing\footnote{\url{https://www.chesspublishing.com}}.
Because the collected annotations vary in quality and factual accuracy, we apply regex- and LLM-based filtering and discard annotations that contradict engine evaluations (e.g., labeling a blunder as a good move).
From about 25k collected examples, 10k remain after this stage.
We then decompose each annotation into \textit{atoms}, each capturing a single logical unit of observation or opinion about a move, such as a tactical consequence, positional assessment, or comparison to an alternative.
We discard annotations containing only a single atom, resulting in 5k training examples.
Finally, an LLM rewrites the atoms into fluent, self-contained commentary.
Each training example contains both the list of atoms in \texttt{<atom>} tags and the resulting commentary in \texttt{<commentary>} tags.

For evaluation, we use 50 examples curated by human players with substantial chess expertise.

\subsection{Tool-augmented LLM judge}
Our tool-augmented LLM judge evaluates chess commentaries in two stages: (1) atomic claim decomposition and (2) tool-augmented verification.

\begin{table}[t]
\centering
\caption{\textbf{Tool suite.} We implement a set of tools that enables the LLM judge to verify the factual correctness of diverse chess claims through chained tool calls.}
\label{tab:tools}
\footnotesize
\vspace{2pt}
\begin{tabular}{lll}
\toprule[1pt]
\textbf{Category} & \textbf{Tool} & \textbf{Purpose} \\
\midrule[0.75pt]

Position Query 
& \texttt{get\_piece\_at}(fen, sq) 
& Return the piece on a given square \\

& \texttt{get\_legal\_moves}(fen) 
& List all legal moves in the position \\

& \texttt{is\_check}(fen) 
& Determine whether the side to move is in check \\

& \texttt{get\_material}(fen) 
& Compute material balance / piece counts \\

& \texttt{get\_squares}(fen, piece, color) 
& Find locations of a piece type \\

\midrule[0.25pt]

Attack / Defense 
& \texttt{get\_attacks}(fen, sq) 
& List squares attacked by a piece on a square \\

& \texttt{get\_attackers}(fen, sq) 
& List pieces attacking a square \\

& \texttt{count\_attackers\_defenders}(fen, sq) 
& Count attacking and defending pieces on a square \\

& \texttt{is\_pinned}(fen, sq) 
& Check whether a piece is pinned \\

& \texttt{check\_threat}(fen, move) 
& Verify existence of a tactical or strategic threat \\

\midrule[0.25pt]

Geometry 
& \texttt{check\_ray\_alignment}(sq\_a, sq\_b) 
& Test alignment along ranks, files, or diagonals \\

& \texttt{get\_squares}(fen, piece, color) 
& Retrieve squares occupied by specific pieces \\

& \texttt{check\_open\_file}(fen, file, color) 
& Check if a file is half-open \\

\midrule[0.25pt]

Variations 
& \texttt{try\_variation}(fen, moves) 
& Simulate a move sequence  \\

\midrule[0.25pt]

Engine Analysis 
& \texttt{eval\_move}(fen, move) 
& Evaluate move quality (e.g., win probability loss) \\

& \texttt{get\_engine\_eval}(fen, move) 
& Get Stockfish evaluation of the move and best move \\

& \texttt{get\_top\_moves}(fen, n) 
& Return top-$n$ engine-recommended moves \\

& \texttt{compare\_moves}(fen, m\_a, m\_b) 
& Compare relative quality of two moves \\

\bottomrule[1pt]
\end{tabular}
\end{table}

\paragraph{Stage 1: Atomic claim decomposition.}
Each generated commentary is decomposed into \textit{candidate atoms}, where each atom represents a position-specific factual statement.
To enable fine-grained verification, each atom is further decomposed into one or more verifiable sub-claims, corresponding to individual facts that can be checked independently.
For example, the atom ``Qe4 centralizes the queen and attacks the bishop on b4'' is decomposed into two sub-claims: (1) ``Qe4 centralizes the queen'' and (2) ``Qe4 attacks the bishop on b4''.
We additionally assign each sub-claim a set of tags (e.g., \textit{quality}, \textit{positional}, \textit{move continuation}) using a lightweight classifier.
These tags provide task-specific guidance for the verification.

\paragraph{Stage 2: Tool-augmented verification.}

\begin{tcolorbox}[title={\textbf{Multi-Tool Verification Example}}, colback=gray!5, colframe=black!50, breakable, fonttitle=\bfseries\small, fontupper=\small, float=t]
\textbf{Generated commentary:} ``Qe4 centralizes the queen and creates a dual threat: it attacks the bishop on b4 and prepares to infiltrate
Black's position via the e-file.''                        
\medskip

\textbf{Candidate atom 1:} Qe4 centralizes the queen and creates a dual threat: it attacks the bishop on b4 and prepares to infiltrate Black's position via the e-file. [{\color{red}\texttimes} Incorrect - 1/3 sub-claims verified]

\begin{itemize}[leftmargin=*, itemsep=2pt]
  \item[{\color{green!60!black}$\checkmark$}] ``Qe4 centralizes the queen'' -- \texttt{get\_piece\_at}(``e4''), \texttt{get\_squares}(``queen''), confirm queen moved from b1 to central e4. \textbf{Verified.}
  \item[{\color{red}\texttimes}] ``Qe4 attacks the bishop on b4'' -- \texttt{get\_attacks}(``e4'') → \texttt{[``b1'', ``c2'', ``c4'', ``d3'', ...]} does not include b4. \textbf{Incorrect.}
  \item[{\color{red}\texttimes}] ``Qe4 prepares to infiltrate via the e-file'' -- \texttt{get\_attacks}(``e4'') shows e-file blocked by Black's e5 pawn; no penetration possible. \textbf{Incorrect.}
\end{itemize}

\textbf{Result:} Tool-based verification catches hallucinations (false claims about attacking b4 and e-file infiltration) that pure LLM
evaluation would miss.
\end{tcolorbox}

Each sub-claim is verified independently by the judge LLM using specialized prompts conditioned on the extracted tags, with mandatory access to chess tools.
Verification is grounded in deterministic tool outputs rather than relying solely on the judge LLM's internal chess knowledge.
A candidate atom is marked \textit{correct} only if all of its sub-claims are verified as true; otherwise, it is marked incorrect.
We implement 18 specialized tools covering positional queries, attack and defense relations, board geometry, variation legality and resulting positions, and engine-based analysis (Table~\ref{tab:tools}).
All tools operate on FEN positions and return structured JSON outputs, enabling fine-grained verification of complex claims.

\subsection{Error analysis}
\label{sec:error_analysis}

\begin{figure}[ht]
    \centering
    \includegraphics[width=0.75\linewidth]{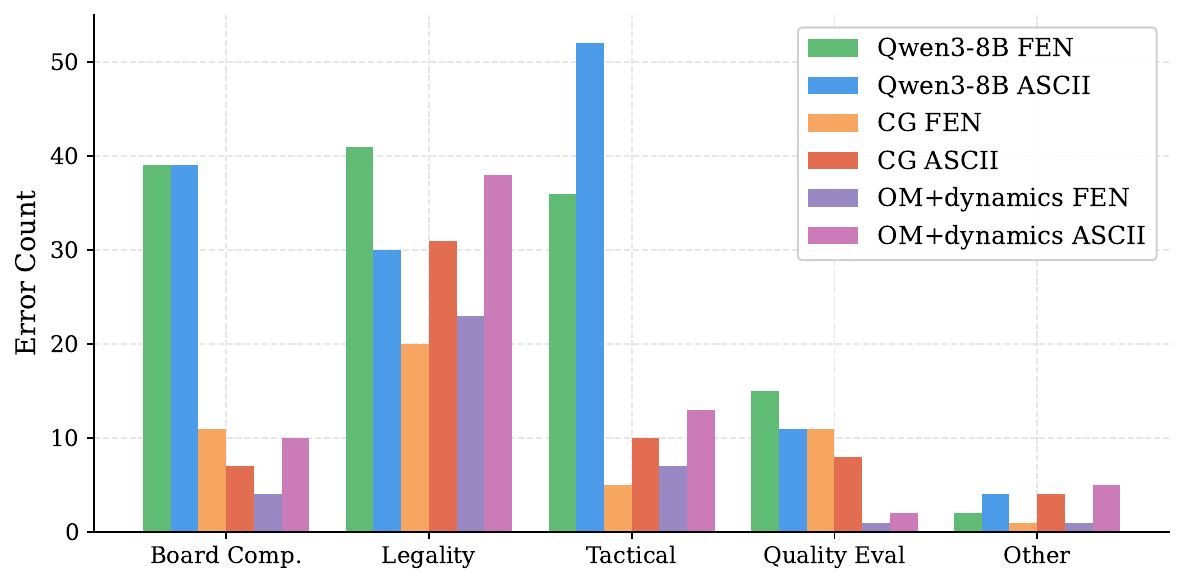}
    \caption{\textbf{Error analysis for commentary generation.} Distribution of error types across six models. Board comprehension errors reflect failures in static board parsing; Legality errors reflect failures in multi-step board state tracking; Tactical errors are those regarding downstream consequences of piece localization; Quality errors reflect failures in position understanding.}
    \label{fig:error-analysis}
\end{figure}

To understand the sources of factual correctness improvements in Table~\ref{table:commentary-table}, we analyze the distribution of error types across training conditions (Figure~\ref{fig:error-analysis}).
We consider the following error categories:
\begin{itemize}[leftmargin=7mm]
\item \textbf{Board comprehension errors}:  
failures to accurately identify the static board configuration, including piece identity, location, and structural properties such as open or blocked files.
\item \textbf{Legality errors}:  
failures to track board state across moves, resulting in illegal sequences.
\item \textbf{Tactical reasoning errors}:  
incorrect reasoning about dynamic consequences, such as asserting nonexistent threats or claiming that a move is forced when legal alternatives exist.
\item \textbf{Quality evaluation errors}:  
incorrect assessments of move strength or positional advantage, including claims that contradict engine analysis or invalid comparisons between positions.
\end{itemize}

We highlight two observations.
First, the base model exhibits a high rate of board comprehension errors in both representations, along with downstream legality and tactical reasoning errors.
Commentary generation fine-tuning substantially reduces board comprehension and tactical reasoning errors, indicating improved static board understanding across representations and auxiliary supervision conditions.
Second, ASCII-aligned auxiliary supervision increases the frequency of illegal move sequences relative to the base model.
This shift coincides with more extensive move continuations in generated commentary, suggesting that auxiliary supervision may encourage more detailed but harder-to-maintain multi-step reasoning.
Thus, while auxiliary supervision improves factual reasoning overall, tracking board state across longer move sequences remains a limitation.

\section{Reasoning examples}
\label{app:sft-samples}

\subsection{Optimal-move prediction}

\begin{tcolorbox}[title={\textbf{Optimal-move prediction}}, colback=gray!5, colframe=black!50, breakable, fonttitle=\bfseries\small]
\begin{lstlisting}
<think>
The current position is: 2rqr1k1/pp1bppb1/3p1np1/4n2p/3NP2P/1BN1BP2/PPPQ2P1/1K1R3R w - -.
It is White's turn to move.

Let me analyze White's candidate moves:

First, consider Bh6.
If played, a natural continuation might be:
Bh6
... Nc4
Bxc4
... Rxc4
Bxg7
Overall, White comes out with a slight edge (about +0.46 pawns).
In this variation, White's win rate is about 54.22%

Another possible choice is g4.
If played, a natural continuation might be:
g4
... hxg4
h5
... Nxh5
Bh6
Overall, White comes out with a slight edge (about +0.73 pawns).
In this variation, White's win rate is about 56.68%

Based on the above analysis, the best move is g4.
It gives White a 56.68%
</think>

<answer>
g4
</answer>
\end{lstlisting}
\end{tcolorbox}

\subsection{Surface task}

\begin{tcolorbox}[title={\textbf{Surface task}}, colback=gray!5, colframe=black!50, breakable, fonttitle=\bfseries\small]
\begin{lstlisting}
<think>
The board in FEN is:
8/3k1PK1/7R/8/8/8/4r3/8 w - -

Output the rank field for rank 2 exactly as written, with no other text.

Rank 2 is the 7th rank field in the board string.
Reading it off gives: 4r3
</think>

<answer>
4r3
</answer>
\end{lstlisting}
\end{tcolorbox}

\subsection{Static state task}

\begin{tcolorbox}[title={\textbf{Static state task}}, colback=gray!5, colframe=black!50, breakable, fonttitle=\bfseries\small]
\begin{lstlisting}
<think>
The board in FEN is:
2rqr1k1/pp1bppb1/3p1np1/4n2p/3NP2P/1BN1BP2/PPPQ2P1/1K1R3R w - -

Using the standard 1/3/3/5/9 piece values and ignoring kings, which side has more material, and by how much?

White material: 38. Black material: 38.
The material is equal.
</think>

<answer>
equal
</answer>
\end{lstlisting}
\end{tcolorbox}

\subsection{Next-state prediction}

\begin{tcolorbox}[title={\textbf{FEN $\to$ FEN (short reasoning)}}, colback=gray!5, colframe=black!50, breakable, fonttitle=\bfseries\small]
\begin{lstlisting}
<think>
I need to apply the move Nxe5 to the position
r2qkbnr/ppp1pppp/8/3Pn3/6b1/2N2N2/PPP2PPP/R1BQKB1R w KQkq -.

Current position analysis:
- It is White's turn.
- The move Nxe5 moves the knight from f3 to e5.
- This captures the knight on e5.

After applying the move:
- The piece layout becomes:
  r2qkbnr/ppp1pppp/8/3PN3/6b1/2N5/PPP2PPP/R1BQKB1R
- It is now Black's turn (b)
- Castling rights: KQkq
- En passant square: -

The resulting board position is:
r2qkbnr/ppp1pppp/8/3PN3/6b1/2N5/PPP2PPP/R1BQKB1R
</think>

<answer>
r2qkbnr/ppp1pppp/8/3PN3/6b1/2N5/PPP2PPP/R1BQKB1R
</answer>
\end{lstlisting}
\end{tcolorbox}

\begin{tcolorbox}[title={\textbf{FEN $\to$ FEN (long reasoning)}}, colback=gray!5, colframe=black!50, breakable, fonttitle=\bfseries\small]
\begin{lstlisting}
<think>
I need to apply the move Nxe5 to the position
r2qkbnr/ppp1pppp/8/3Pn3/6b1/2N2N2/PPP2PPP/R1BQKB1R w KQkq -.

Current position analysis:
- It is White's turn.
- The position has 30 pieces on the board.

Current FEN position rank by rank:
Rank 8: r2qkbnr -- 6 piece(s): r, q, k, b, n, r
Rank 7: ppp1pppp -- 7 piece(s): p, p, p, p, p, p, p
Rank 6: 8 -- empty rank
Rank 5: 3Pn3 -- 2 piece(s): P, n
Rank 4: 6b1 -- 1 piece(s): b
Rank 3: 2N2N2 -- 2 piece(s): N, N
Rank 2: PPP2PPP -- 6 piece(s): P, P, P, P, P, P
Rank 1: R1BQKB1R -- 6 piece(s): R, B, Q, K, B, R

Move analysis:
- The White knight on f3 moves to e5.
- Source square f3 (rank 3, file f) will be emptied.
- Destination square e5 (rank 5, file e) will be occupied.
- This captures the black knight on e5.

Affected ranks: rank 5, rank 3
- Rank 5: 3Pn3 -> 3PN3 (e5 changes from n to N)
- Rank 3: 2N2N2 -> 2N5 (N removed from f3)

Resulting position verification:
Rank 8: r2qkbnr
Rank 7: ppp1pppp
Rank 6: 8
Rank 5: 3PN3 [changed]
Rank 4: 6b1
Rank 3: 2N5 [changed]
Rank 2: PPP2PPP
Rank 1: R1BQKB1R

After applying the move:
- The piece layout becomes:
  r2qkbnr/ppp1pppp/8/3PN3/6b1/2N5/PPP2PPP/R1BQKB1R

The resulting board position is:
r2qkbnr/ppp1pppp/8/3PN3/6b1/2N5/PPP2PPP/R1BQKB1R
</think>

<answer>
r2qkbnr/ppp1pppp/8/3PN3/6b1/2N5/PPP2PPP/R1BQKB1R
</answer>
\end{lstlisting}
\end{tcolorbox}

\begin{tcolorbox}[title={\textbf{ASCII $\to$ ASCII (short reasoning)}}, colback=gray!5, colframe=black!50, breakable, fonttitle=\bfseries\small]
\begin{lstlisting}
<think>
I need to apply Nxe5 to the current position.
It is White's turn.

Current board:
r . . q k b n r
p p p . p p p p
. . . . . . . .
. . . P n . . .
. . . . . . b .
. . N . . N . .
P P P . . P P P
R . B Q K B . R

Move Nxe5: The knight moves from f3 to e5.
It captures the knight on e5.

Updated board:
r . . q k b n r
p p p . p p p p
. . . . . . . .
. . . P N . . .
. . . . . . b .
. . N . . . . .
P P P . . P P P
R . B Q K B . R

The resulting board position is shown above.
</think>

<answer>
r . . q k b n r
p p p . p p p p
. . . . . . . .
. . . P N . . .
. . . . . . b .
. . N . . . . .
P P P . . P P P
R . B Q K B . R
</answer>
\end{lstlisting}
\end{tcolorbox}

\begin{tcolorbox}[title={\textbf{ASCII $\to$ ASCII (long reasoning)}}, colback=gray!5, colframe=black!50, breakable, fonttitle=\bfseries\small]
\begin{lstlisting}
<think>
I need to apply the move Nxe5 to the current position.

Current board:
r . . q k b n r
p p p . p p p p
. . . . . . . .
. . . P n . . .
. . . . . . b .
. . N . . N . .
P P P . . P P P
R . B Q K B . R

It is White's turn.
There are 30 pieces on the board.

Pieces on affected ranks:
Rank 5: white pawn on d5, black knight on e5
Rank 3: white knight on c3, white knight on f3

Move analysis:
- Nxe5: White knight (N) on f3 moves to e5.
- This captures the black knight on e5.
- Square f3: N -> .
- Square e5: captured -> N

Affected ranks: 5, 3
- Rank 5: e5: n -> N
  before: . . . P n . . .
  after:  . . . P N . . .
- Rank 3: f3: N -> .
  before: . . N . . N . .
  after:  . . N . . . . .

Board comparison (before -> after):
Rank 8: r . . q k b n r  (unchanged)
Rank 7: p p p . p p p p  (unchanged)
Rank 6: . . . . . . . .  (unchanged)
Rank 5: . . . P n . . .  ->  . . . P N . . . [changed]
Rank 4: . . . . . . b .  (unchanged)
Rank 3: . . N . . N . .  ->  . . N . . . . . [changed]
Rank 2: P P P . . P P P  (unchanged)
Rank 1: R . B Q K B . R  (unchanged)

Updated board:
r . . q k b n r
p p p . p p p p
. . . . . . . .
. . . P N . . .
. . . . . . b .
. . N . . . . .
P P P . . P P P
R . B Q K B . R
</think>

<answer>
r . . q k b n r
p p p . p p p p
. . . . . . . .
. . . P N . . .
. . . . . . b .
. . N . . . . .
P P P . . P P P
R . B Q K B . R
</answer>
\end{lstlisting}
\end{tcolorbox}

\begin{tcolorbox}[title={\textbf{FEN $\to$ ASCII $\to$ FEN (short reasoning)}}, colback=gray!5, colframe=black!50, breakable, fonttitle=\bfseries\small]
\begin{lstlisting}
<think>
I need to apply Nxe5 to
r2qkbnr/ppp1pppp/8/3Pn3/6b1/2N2N2/PPP2PPP/R1BQKB1R w KQkq -.
Nxe5: The knight moves from f3 to e5.
It captures the knight on e5.

Updated board in ASCII:
r . . q k b n r
p p p . p p p p
. . . . . . . .
. . . P N . . .
. . . . . . b .
. . N . . . . .
P P P . . P P P
R . B Q K B . R

Converting changed ranks back to FEN:
  Rank 5: . . . P N . . . -> 3PN3
  Rank 3: . . N . . . . . -> 2N5

Resulting FEN:
r2qkbnr/ppp1pppp/8/3PN3/6b1/2N5/PPP2PPP/R1BQKB1R
</think>

<answer>
r2qkbnr/ppp1pppp/8/3PN3/6b1/2N5/PPP2PPP/R1BQKB1R
</answer>
\end{lstlisting}
\end{tcolorbox}

\begin{tcolorbox}[title={\textbf{FEN $\to$ ASCII $\to$ FEN (long reasoning)}}, colback=gray!5, colframe=black!50, breakable, fonttitle=\bfseries\small]
\begin{lstlisting}
<think>
I need to apply the move Nxe5 to the position
r2qkbnr/ppp1pppp/8/3Pn3/6b1/2N2N2/PPP2PPP/R1BQKB1R w KQkq -.

First, let me convert the FEN position to an ASCII board.
r2qkbnr/ppp1pppp/8/3Pn3/6b1/2N2N2/PPP2PPP/R1BQKB1R
converted rank by rank:
Rank 8: r2qkbnr -> r . . q k b n r
Rank 7: ppp1pppp -> p p p . p p p p
Rank 6: 8 -> . . . . . . . .
Rank 5: 3Pn3 -> . . . P n . . .
Rank 4: 6b1 -> . . . . . . b .
Rank 3: 2N2N2 -> . . N . . N . .
Rank 2: PPP2PPP -> P P P . . P P P
Rank 1: R1BQKB1R -> R . B Q K B . R

Current board:
r . . q k b n r
p p p . p p p p
. . . . . . . .
. . . P n . . .
. . . . . . b .
. . N . . N . .
P P P . . P P P
R . B Q K B . R

Applying Nxe5: the knight moves from f3 to e5.
It captures the black knight on e5.

Updated board:
r . . q k b n r
p p p . p p p p
. . . . . . . .
. . . P N . . .
. . . . . . b .
. . N . . . . .
P P P . . P P P
R . B Q K B . R

Now converting the ASCII board back to FEN:
Rank 8: r . . q k b n r -> r2qkbnr
Rank 7: p p p . p p p p -> ppp1pppp
Rank 6: . . . . . . . . -> 8
Rank 5: . . . P N . . . -> 3PN3
Rank 4: . . . . . . b . -> 6b1
Rank 3: . . N . . . . . -> 2N5
Rank 2: P P P . . P P P -> PPP2PPP
Rank 1: R . B Q K B . R -> R1BQKB1R

The resulting position:
r2qkbnr/ppp1pppp/8/3PN3/6b1/2N5/PPP2PPP/R1BQKB1R
</think>

<answer>
r2qkbnr/ppp1pppp/8/3PN3/6b1/2N5/PPP2PPP/R1BQKB1R
</answer>
\end{lstlisting}
\end{tcolorbox}

\subsection{Legal-move prediction}

\begin{tcolorbox}[title={\textbf{Legal-move prediction}}, colback=gray!5, colframe=black!50, breakable, fonttitle=\bfseries\small]
\begin{lstlisting}[escapeinside={*}{*}]
<think>
White pieces - scan each rank:
  Rank 4: 3NP2P. White: N(d4), P(e4), P(h4).
  Rank 3: 1BN1BP2. White: B(b3), N(c3), B(e3), P(f3).
  Rank 2: PPPQ2P1. White: P(a2), P(b2), P(c2), Q(d2), P(g2).
  Rank 1: 1K1R3R. White: K(b1), R(d1), R(h1).
White has: Knight(d4), Pawn(e4), Pawn(h4), Bishop(b3), Knight(c3), Bishop(e3), Pawn(f3), Pawn(a2), Pawn(b2), Pawn(c2), Queen(d2), Pawn(g2), King(b1), Rook(d1), Rook(h1).

Consider Bishop on e3.
Bishops move diagonally.
  f4(*$\checkmark$*), g5(*$\checkmark$*), h6(*$\checkmark$*).
  d4(own knight, stop).
  f2(*$\checkmark$*), g1(*$\checkmark$*).
  d2(own queen, stop).

Select Bf4 - improves bishop activity.
</think>

<answer>Bf4</answer>
\end{lstlisting}
\end{tcolorbox}

\subsection{Representation mapping}

\begin{tcolorbox}[title={\textbf{FEN-to-ASCII mapping}}, colback=gray!5, colframe=black!50, breakable, fonttitle=\bfseries\small]
\begin{lstlisting}[escapeinside={*}{*}]
<think>
I need to convert the FEN position to ASCII format.
The FEN position is: 2rqr1k1/pp1bppb1/3p1np1/4n2p/3NP2P/1BN1BP2/PPPQ2P1/1K1R3R

In FEN notation:
- Uppercase letters represent white pieces (K, Q, R, B, N, P)
- Lowercase letters represent black pieces (k, q, r, b, n, p)
- Numbers represent consecutive empty squares

Converting rank by rank (from rank 8 to rank 1):
Rank 8: 2rqr1k1 *$\rightarrow$* . . r q r . k .
Rank 7: pp1bppb1 *$\rightarrow$* p p . b p p b .
Rank 6: 3p1np1 *$\rightarrow$* . . . p . n p .
Rank 5: 4n2p *$\rightarrow$* . . . . n . . p
Rank 4: 3NP2P *$\rightarrow$* . . . N P . . P
Rank 3: 1BN1BP2 *$\rightarrow$* . B N . B P . .
Rank 2: PPPQ2P1 *$\rightarrow$* P P P Q . . P .
Rank 1: 1K1R3R *$\rightarrow$* . K . R . . . R

The complete ASCII representation:
</think>

<answer>
. . r q r . k .
p p . b p p b .
. . . p . n p .
. . . . n . . p
. . . N P . . P
. B N . B P . .
P P P Q . . P .
. K . R . . . R
</answer>
\end{lstlisting}
\end{tcolorbox}

\section{Limitations}
\label{app:limitations}

While our results demonstrate the effectiveness of representation-aligned auxiliary supervision for domain adaptation, several limitations remain.
First, our study focuses on chess. %
Although this setting allows us to isolate the effect of representation choice from the training objective, it remains to evaluate how well our findings generalize to less structured settings.
Second, we operationalize representation compatibility using empirical proxies such as task performance and reasoning behavior.
While these provide useful signals, they do not constitute a formal measure. %
Developing more principled measures of compatibility, e.g., based on internal representations, could strengthen the framework and improve its applicability across domains.
Finally, our framework assumes access to auxiliary supervision derived from environment structure.
In domains lacking such structure, applying our approach may require additional modeling effort.
Determining how to construct useful auxiliary supervision in such settings is an important direction for future work.

\section{Broader impacts}
\label{app:broader-impacts}

This work studies how representation choices affect domain adaptation in language models and how environment-derived auxiliary supervision can be aligned with those representations.
More effective adaptation methods may facilitate the development of models for specialized domains, including settings where reliable structured reasoning is important.
At the same time, improved adaptation does not guarantee robustness or factual reliability, particularly when auxiliary supervision or the underlying environment is imperfectly specified.
We view this work as a step toward more principled language model adaptation and emphasize careful evaluation before applying such methods to consequential tasks.

\end{document}